# Bombardier Beetle Optimizer: A Novel Bio-Inspired Algorithm for Global Optimization

Hisham A. Shehadeh[1,*] . Mohd Yamani Idna Idris[2] . Iqbal H. Jebril[3]

[1,*]Department of Computer Sciences, Faculty of Information Technology and Computer Science, Yarmouk University, Irbid 21163, Jordan. H.shehadeh@yu.edu.jo

https://orcid.org/0000-0003-4054-6561

[2]Department of Computer System and Technology, Faculty of Computer Science and Information Technology, University of Malaya, Kuala Lumpur, Malaysia, yamani@um.edu.my

https://orcid.org/0000-0003-4894-0838

[3]Department of Mathematics, Faculty of Science and Information Technology, Al-Zaytoonah University of Jordan, Amman, Jordan. i.jebril@zuj.edu.jo

https://orcid.org/0000-0003-4348-6197

[*]Corresponding Author (Hisham A. Shehadeh)

**Abstract.**

In this paper, a novel bio-inspired optimization algorithm is proposed, called "Bombardier Beetle Optimizer (BBO)". This type of species is very intelligent, which has an ability to defense and escape from predators. The principles of the former one is inspired by the defense mechanism of Bombardier Beetle against the predators, which the Bombardier Beetle triggers a toxic chemical spray when it feels threatened. This reaction occurs in a specialized reaction chamber inside its abdomen and includes a well regulated enzymatic mechanism, which comprises hot water vapor, oxygen, and irritating substances like p-benzoquinones. In addition, the proposed BBO simulates also the escape mechanism of Bombardier Beetle from predator, which it has the ability to calculate its distance from predator and it can fly away. The BBO is tested with optimizing "Congress on Evolutionary Computation (CEC 2017)" test bed suites. Moreover, it is compared against well-known metaheuristic optimization algorithms includes "Chernobyl Disaster Optimizer (CDO)", "Grey Wolf Optimizer (GWO)", "Particle Swarm Optimization (PSO)" Bermuda Triangle Optimizer (BTO), "Sperm Swarm Optimization (SSO)" and "Gravitational Search Algorithm (GSA)". The outcomes of this paper prove the BBO's efficiency in which outperforms the other algorithms in terms of convergence rate and quality of results.

## 1 Introduction

Real life problems have been increased with increase life demands. Example of these problems are Wireless Sensor Network (WSN) problem, biomedical problems, engineering design problems, etc. These problems can be converted to mathematical models, however, it is a challenging task to solve these problems efficiently, especially when they have a set of local minima in their geometry [1].

To tackle these problems, we rely on optimization algorithms, which are systematic methods and high level procedures designed to search for optimal or near-optimal solutions [2, 3]. These algorithms are inspired by diverse natural and artificial systems, such as biology, social behavior, physics and even historical events, which are making them versatile and robust for real-world applications. These algorithms are broadly categorized into four main classes, including, physical based class, swarm and bioinspired based class, evolutionary based class, and manmade based class. Examples of physical based class are Chernobyl Disaster Optimizer (CDO) [4] and Bermuda Triangle Optimizer (BTO) [5]. Examples of swarm based and bioinspired class are Particle Swarm Optimization (PSO) [6], Grey Wolf Optimizer (GWO) [7] and Sperm Swarm Optimization (SSO) [8,9]. An example of evolutionary based class is Genetics Algorithm (GA) [10,11]. Examples of manmade class are Five Phases Algorithm (FFA) [12] and Fireworks Algorithm (FA) [13].

The advantages of the GSA, SSO, PSO, CDO, BTO, and GWO algorithms can be listed as follows [4-9]:

- SSO and PSO are simple to utilize and understand.

- Usually, PSO simply required to be tuned for the social coefficient, inertia weight, and cognitive coefficient. This simplicity in adjusting of parameters helps mitigate error appearance in code.
- SSO uses random parameters in its operation, such as temperature value, and pH value, which are an exact values of vital signs of female reproduction system. To mitigate gaps between these tuning parameters SSO utilizes the logarithm to normalize them and helps reduce error appearance in code.
- PSO and SSO converge faster in various kinds of problems.
- SSO, GWO and PSO can be utilized to solve different kinds of optimization problems, such as both discrete and continuous problems, solving multiple objectives optimization problems, and solving problems without and with constraints.
- SSO's adaptive strategies provide better convergence than existing classical approaches, which have trouble with high-dimensional restrictions.
- The novel inspiration for CDO and BTO comes from a distinctive and significant historical and real-world occurrence that provides new insights into optimization methods.
- The designs of CDO and BTO allow for effective exploration of the solution space domain of any issue by avoiding local optima and striking a balance between exploration and exploitation tactics.
- CDO is a strong and effective method for resolving challenging optimization issues.
- CDO employs random tuning parameter values, which correspond to the precise values of the alpha, beta, and gamma particles spreading speeds as well as the human walking pace. Additionally, it applies the logarithm to these data in order to normalize them.
- To avoid local minima, BTO also uses random variables in its process.

However, they are facing various problems, such as tripping in local minima, slow convergence, and bad local searching ability [2,4,5,8,9,14]. For these reasons, we are motivated to propose new optimization algorithm that mitigate these problems. This optimizer called "Bombardier Beetle Optimizer (BBO)", which inspired by the defense mechanism of Bombardier Beetle against the predators.

The rest of the paper is listed as follows. Literature reviews are discussed in Section 2. The Bombardier Beetle behavior, its chemical reactions and Bombardier Beetle Optimizer are discussed in Sec. 3. Experimental and outcomes of the proposed method are presented in Sec.4. Discussion is previewed in Section 5. We conclude the outcomes in Sec.6.

## 2 Literature review

The goal of this section is to review on the prior well-known categories to compare them later on against the proposed BBO algorithm. Hence, "Chernobyl Disaster Optimizer (CDO)", "Bermuda Triangle Optimizer (BTO)", "Particle Swarm Optimization (PSO)", "Grey Wolf Optimizer (GWO)" and "Sperm Swarm Optimization (SSO)" are selected in this work based on their aforementioned advantages.

### 2.1 "Standard Sperm Swarm Optimization (SSO)"

Sperm Swarm Optimization (SSO) is a novel optimization approach that proposed by Shehadeh et al. [8,9] SSO mimics the swarm of sperm swimming to reach the egg during fertilization procedure. Fig. 1 illustrates this procedure.

This algorithm utilizes three swarm velocities, which are the beginning velocity, current velocity, and global velocity. The sperm that is closest to the egg wins the global one. This velocity is depicted in Fig. 2 [8]. These velocities are shown in the following mathematical models. The previously mentioned velocities are affected by the pH and temperature, which are random values between 7 and 14 and 35.1 and 38.5, respectively.

$$Initial_{Velocity} = D.V_i.Log_{10}(pH_{Rand_1}) \quad (1)$$

$$Current\_Best\_Solution = Log\_10(pH - Rand_2) \cdot Log_{10}(Temp\_Rand).(sb\_solution[] - current[]) \quad (2)$$

$$Global\_Best\_Solution = Log\_10(pH - Rand_3) \cdot Log_{10}(Temp\_Rand_2).(sgb\_solution[] - current[]) \quad (3)$$

where D is a velocity damping factor. Random values between zero and one are assigned to this component. The values of $pH_{Rand_1}$, $pH_{Rand_2}$, and $pH_{Rand_3}$ are random numbers that range from seven to fourteen. The temperature values, $Temp\_Rand_1$ and $Temp\_Rand_2$, pick a random value between 35.1 and 38.5. As of right now, the best solution is $sb\_solution[]$. The winner's global best solution is denoted by $sgb\_solution[]$.

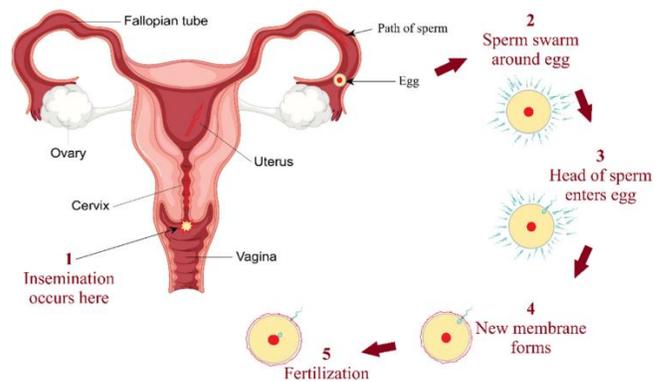

**Fig 1.** Fertilization process [8]

The following equation can be used to combine these velocities. where $v[]$ is the SSO velocity rule.

$$v[] = Initial_{Velocity} + Current\_Best\_solution + Global\_Best\_Solition \qquad (4)$$

The current best solution can be presented in the following formula.

$$Current[] = Current[] + v[] \qquad (5)$$

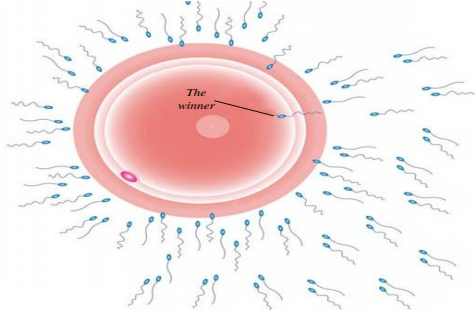

**Fig 2.** Swarm of sperm and the winner [8]

## 2.2 "Standard Chernobyl Disaster Optimizer (CDO)"

The "Chernobyl Disaster Optimizer (CDO)" was inspired by the 1986 Chernobyl nuclear accident. This technique is proposed by Shehadeh in 2023 [4] which simulates the effects and transmission of radiation particles that target humans in order to get around optimization problems. The optimizer takes into account alpha (α), beta (β), and gamma (γ) radiation particles. These particles and the explosion zone are depicted in Fig. 3 [4].

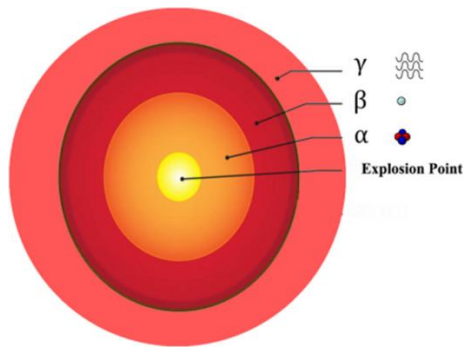

Fig 3. The radiation particles and explosion point [4]

Shehadeh makes the assumption that the gamma, beta, and alpha particles are now located at are $X_\gamma(t)$, $X_\beta(t)$, and $X_\alpha(t)$, respectively, in CDO. The propagation of the gamma, beta, and alpha particles is given by the following models, in that order:

$$\rho_\gamma = \frac{x_h}{S_\gamma} - (WS_h \cdot rand()) \qquad (6)$$

$$\rho_\beta = \frac{x_h}{0.5 \cdot S_\beta} - (WS_h \cdot rand()) \qquad (7)$$

$$\rho_\alpha = \frac{x_h}{0.25 \cdot S_\alpha} - (WS_h \cdot rand()) \qquad (8)$$
$$x_h = r^2 \cdot \pi \qquad (9)$$
$$S_\gamma = \log(rand(1:300,000)) \qquad (10)$$
$$S_\beta = \log(rand(1:270,000)) \qquad (11)$$
$$S_\alpha = \log(rand(1:160,000)) \qquad (12)$$
$$WS_h = 3 - 1 * ((3)/\text{Maximum\_Iteration}) \qquad (13)$$

where $x_h$ is the region where a person would walk within a circle with a random radius of 0 to 1. The normalized and random gamma, beta, and alpha particle speeds are $S_\gamma$, $S_\beta$, and $S_\alpha$. According to Equation (13), $WS_h$ or the human walking speed, decreases linearly from 3 km to 0. The following formulae can be used to determine the difference between the locations of the gamma, beta, and alpha particles and the total position, respectively:

$$\Delta_\gamma = |A_\gamma \cdot X_\gamma(t) - X_T(t)| \qquad (14)$$
$$\Delta_\beta = |A_\beta \cdot X_\beta(t) - X_T(t)| \qquad (15)$$
$$\Delta_\alpha = |A_\alpha \cdot X_\alpha(t) - X_T(t)| \qquad (16)$$

where the area of a circle with a random radius between 0 and 1, as shown in Figure 3., $A_\gamma$, $A_\beta$, and $A_\alpha$ represent the propagation areas of the gamma, beta, and alpha particles, respectively.

$$A_\gamma = A_\beta = A_\alpha = r^2 \cdot \pi \qquad (17)$$

Where $X_T$ is the average speeds of all particles.
$$X_T = \frac{v_\gamma + v_\beta + v_\alpha}{3} \qquad (18)$$

The gamma, beta, and alpha particles' Gradient Descent Factors, denoted by $v_\gamma$, $v_\beta$, and $v_\alpha$, respectively, which are computed using the following formulas to get the best answer [4]:

$$v_\gamma = (X_\gamma(t) - \rho_\gamma \cdot \Delta_\gamma) \qquad (19)$$
$$v_\beta = 0.5 \cdot (X_\beta(t) - \rho_\beta \cdot \Delta_\beta) \qquad (20)$$
$$v_\alpha = 0.25 \cdot (X_\alpha(t) - \rho_\alpha \cdot \Delta_\alpha) \qquad (21)$$

## 2.3 "Standard Bermuda Triangle Optimizer (BTO)"

Shehadeh [5] proposed this optimizer, as new metaheuristic optimization technique, in 2025. This algorithm mimics the enigmatic Bermuda Triangle phenomenon, when several ships and airplanes are drawn into the Bermuda triangle's region and vanish there. Shehadeh makes the assumption that there are two regions of forces in BTO.

The first one is the exploitation area, which has a strong attraction force and can form a Bermuda triangle. The second one, the circle encircled by the Bermuda triangle, which has a less attraction force. This area can be considered for exploration. These areas are represented in Fig. 4. According to Newton's method of gravity in Eq. (22) each attracted object in BTO takes a random position on the search space domain and a force probability value before being pulled to the triangle's center.

$$G_{force} = \frac{CUG \cdot M_1 \cdot M_2}{r^2} \quad (22)$$

Where the CUG is $6{,}67 \times 10^{-11} \text{Nm}^2\text{Kg}^{-2}$, which is constant of universal gravitation. $M_1$ represents the mass at the center of the Bermuda Triangle that generates the gravitational field, modeled as a random value. $M_2$ represents the mass that affect by $M_1$, modeled as a random value. $r$ is the distance between $M_1$ and $M_2$, modeled as a random value.

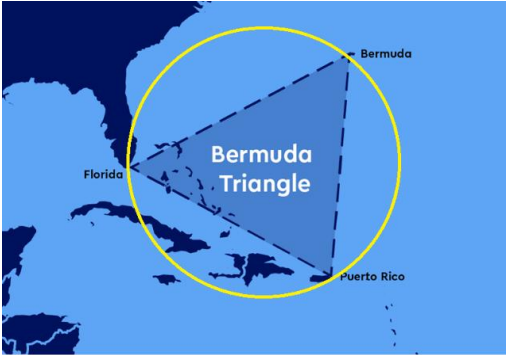

**Fig 4.** Massive attraction force and less attraction force [5]

To replicate the precise movement of things in the ocean, where they will be impacted by tides and ocean flow, the movement of these objects will be modified using the levy and chaos approaches in BTO. Prescience is primarily defined as any object, whether inside or outside the Bermuda Triangle that is randomly assigned a value more than 0.5 or less than 0.5. The algorithm performs a subtraction operation, indicating a strong gravitational force, when this value surpasses 0.5. The region of the Bermuda Triangle, as seen in Fig. 4, is used to compute this force. This enormous attraction already applies to objects situated within the Bermuda Triangle. The likelihood that the alternative hypothesis is correct, which represents the probability that the alternative hypothesis holds true. In Equation (23), the Probability of Force (PoF) is defined as (1 – p-value). On the other hand, the algorithm performs an addition operation, signifying a smaller attractive force, when the prescience value is less than 0.5. As seen in Fig. 4, this force is calculated by deducting the area of the surrounding region (the yellow circular zone) from the area of the Bermuda Triangle. The object moves toward the ideal solution as formulated by Equation (23), based on this calculation [5].

$$X_{i,j}(Iter_i + 1)$$

$$= \begin{cases} choas \times Triangle_{area} \times Acc \times best(x_j) - PoF \\ \times ((UB - LB \times Zone_{BF} + LB), Random > 0.5 \\ choas \times circle_{area} \times Acc \times best(x_j) - PoF \\ \times ((UB - LB \times Zone_{BF} + LB), Otherwise \end{cases} \quad (23)$$

Where $LB$ and $UB$ refer to the lower and upper bounds of the problem's search space. $Zone_{BF}$ denotes the Bermuda Force Zone, which can be coefficient computed using Equation (24). $PoF$ stands for the probability ratio of the Bermuda force, which is derived from Equation (25). $best(x_j)$ represents the best value obtained so far. $A_{cc}$ is the acceleration function, which is used to enhance the flow of the ocean current, and it can be calculated by exponential function as in Equation (26). $Triangle_{area}$ is the area of Bermuda triangle. $circle_{area}$ is the subtraction between triangle area and circle area, which refers to the surrounding area of Bermuda [5].

$$Zone_{BF} = area_{min} + Iter_1 * \left(\frac{area_{max} - area_{min}}{Iter_T}\right) \quad (24)$$

Where $area_{min}$ is the logarithm of the minimum area of the Bermuda force zone, which is 500,000 square miles. The logarithm is used for normalization. $Iter_1$ is the counter value at the $i_{th}$ iteration. $area_{max}$ is the logarithm of the maximum area of the Bermuda force zone, which is 1,510,000 square miles, also normalized using the logarithm. $Iter_T$ is the maximum number of iterations. The Bermuda force probability ratio can be determined using the following equation. In statistical terms, the p-value is the null hypothesis, which is true, while (1 – p-value) indicates the probability that the alternative hypothesis, which is also true, as shown in Equation (24) [5].

$$PoF = 1 - \left[\frac{iter_i\frac{1}{G_{Force}}}{iter_T\frac{1}{G_{Force}}}\right] \quad (25)$$

$$A_{cc} = r \times e^{\left(-20x\left(\frac{Iter_1}{Iter_T}\right)\right)} \quad (26)$$

Where $iter_i$ is the $i_{th}$ iteration counter. $iter_T$ is the maximum number of iterations. $r$ is random value. $G_{Force}$ is Bermuda triangle force, which is Eq. (22) [5].

## 2.4  "Standard Particle Swarm Optimization (PSO)"

Particle Swarm Optimization (PSO) is a swarm-based metaheuristic optimization algorithm, which is inspired by the social behavior of birds while searching food. It was introduced by Kennedy and Eberhart in 1995 [6]. The position of each particle will be adjusted on search space domain based on three steps as in Equation (27), which are initial velocity of particle, best velocity of particle and global best velocity of particle. The global one is the leader of bird swarm. This algorithm has a set of parameters in its velocity rule, which are inertia weight (w), cognitive factor($c_1$), and social factor($c_2$). The $c_1$ and $c_2$ are always have a value of 2 [6].

$$fV_{i,m}^{(t+1)} = w * v_{i,m}^{(t)} + c_1 * rand_1() * \left(pbest_{i,m} - x_{i,m}^{(t)}\right) \\ + c_2 * rand_2() * \left(gbest_m - x_{i,m}^{(t)}\right) \quad (27)$$

## 2.5 "Standard Grey Wolf Optimizer (GWO)"

GWO is proposed by Mirjalili et al. in 2014 [7], which is a nature inspired algorithm that mimics the hunting behavior and social hierarchy of grey wolves in nature. Mirjalili et al. divided these wolfs to a set of sections, such as follows [7]:

1. Alpha (α) part is Leader (best solution);
2. Beta (β) part is second level (second-best solution);
3. Delta (δ) part is third level (third-best solution).

These parts of social hierarchy have a set of strategy, such as searching for prey, which is the exploration part of the algorithm. Attacking prey and hunting, which is exploitation part of the algorithm. At the end he takes the total position of these parts of social hierarchy and divided them on three, as in the following equation [7].

$$total\ Position = \frac{Position_\alpha \times Position_\beta \times Position_\delta}{3} \quad (28)$$

## 3 Chemical Reactions in the Bombardier Beetle

The "Carabidae family" of bombardier beetles has a special and potent chemical defense system. It repels predators with a scorching, toxic chemical spray when it feels threatened. This reaction occurs in a specialized reaction chamber inside its abdomen and includes a well regulated enzymatic mechanism [15-21]. The Bombardier Beetle while activates its chemical defense system is depicted in Fig.5.

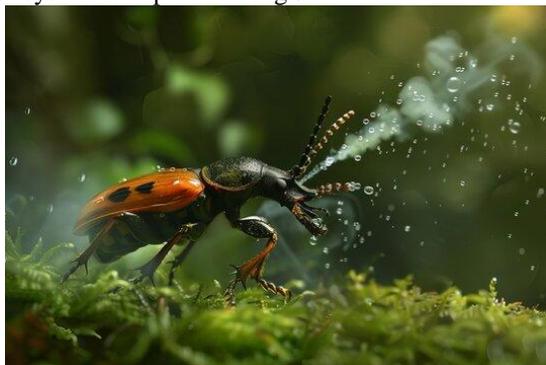

**Fig 5.** The Bombardier Beetle while activates its chemical defense system

### 3.1 *Chemical Reactions*

In its abdomen, the insect stores two distinct chemicals, which are as follows [15-20]:
- Hydroquinone ($C_6H_4(OH)_2$)
- Hydrogen peroxide ($H_2O_2$)

These substances are combined in a reaction chamber with enzymes such as catalase and peroxidase when the beetle is in danger. A hot spray is ejected as a result of an exothermic reaction that is triggered by it. The following chemical reactions are produced as a result of chemical reactions [17]:

1. Hydrogen peroxide decomposition:
   - $2\ H_2O_2 \rightarrow 2\ H_2O + O_2$ (catalyzed by catalase)
2. Hydroquinone oxidation:
   - $C_6H_4(OH)_2 + O_2 \rightarrow C_6H_4O_2$ (quinone) + $H_2O$ (catalyzed by peroxidase)

The outcome of this reaction can be summarized as follows, the mixture is heated to over 100°C as a result of the reaction's quick release of gas and heat. The toxic mixture is forced through a nozzle in bursts by the pressure. Predators are efficiently deterred by the expelled spray, which comprises hot water vapor, oxygen, and irritating substances like p-benzoquinones. Table 1 shows the components of chemical reaction and the outcome [17].

**Table 1** components of chemical reaction and the outcome [17]

| Component | Function |
|---|---|
| Hydroquinone | Fuel, oxidized to form quinones |
| Hydrogen peroxide | Oxidizer, decomposes to release $O_2$ |
| Catalase/Peroxidase | Enzymes that catalyze reactions |
| p-Benzoquinone | Irritant and toxic compound |
| Outcome | Hot, noxious spray (defense mechanism) |

### 3.2 The Bombardier Beetle Optimizer

In this paper, the prior information of defense mechanism of the Bombardier Beetle motivated us to mimics it as optimization algorithm, called Bombardier Beetle Optimizer (BBO). The algorithm steps are organized in the following points.

#### A. *Initialization of BBO:*

This algorithm proposes that there is a set of Bombardier Beetle searching a search space of a problem. Depending on where it falls in the problem-solving space, each one produces a value assignment for the decision variables. Thus, each Bombardier Beetle is a possible solution that can be represented mathematically as a vector. These victors can be mathematically converted as a matrix as depicted in the Fig 6.

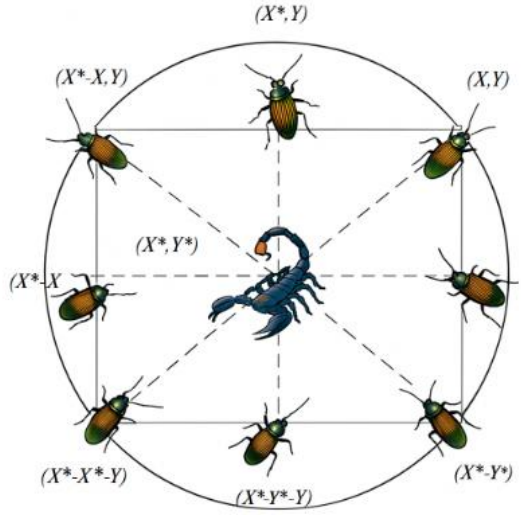

**Fig 6.** Positions of Bombardier Beetle and predator

The initialization of each position of Bombardier Beetle on search space should not exceed the upper bound and not reduce than the lower bound of the problem-solving space. The mathematical equation of initialization can be formularized as follows:

$$x_{i,d} = lb_d + r \cdot (ub_d - lb_d) \qquad (29)$$

### B. Defense mechanism of Bombardier Beetle

In this section, we simulate the defense mechanism of the Bombardier Beetle against the predators. The area of defense is a form of circle, which it triggers a toxic chemical spray when the area of predator intersects the area of Bombardier Beetle. Based on that we have three possible situations as follows:

- The predator is away from the Bombardier Beetle, which means that there is no overlap or intersection area between Bombardier Beetle and predator. In this case the between Bombardier Beetle will not trigger the toxic chemical spray. These areas are depicted in the Fig 7.

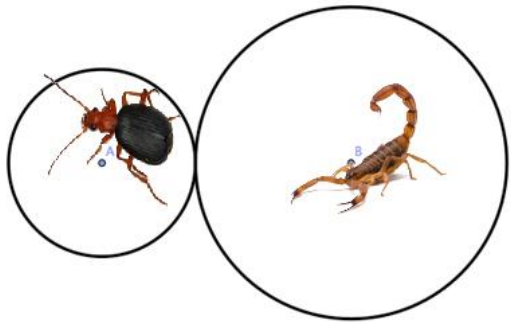

**Fig 7.** The predator is away from the Bombardier Beetle, which is no intersection area between them (area equal to zero).

- The rest two cases are as follows:
  a. The area of predator intersects the area of Bombardier Beetle, which is depicted in Fig. 8 (a).
  b. The area of predator overlaps the area of Bombardier Beetle, which is depicted in Fig. 8 (b).

In these two cases, the Bombardier Beetle will activate the defense mechanism by triggering the toxic chemical spray to the predator. These two cases are depicted in figure 8 (a), and (b).

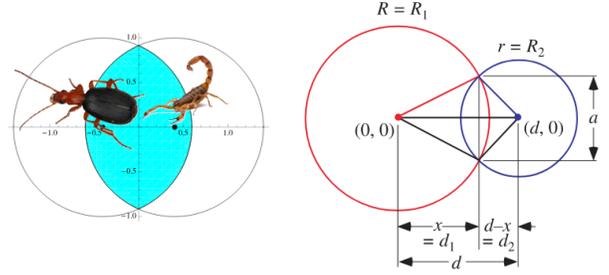

(a)

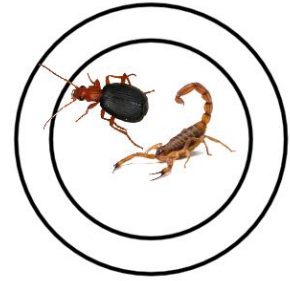

(b)

**Fig 8.** The area of predator intersects with the area of Bombardier Beetle in (a). The area of predator overlaps with the area of Bombardier Beetle in (b).

Mathematically speaking, the following equation is used to calculate the intersection area of two circle, which is the area of spraying the toxic chemical spray.

$$\begin{aligned} A &= A(R, d_1) + A(r, d_2) \qquad (30) \\ &= r^2 \cos^{-1}\left(\frac{d^2 + r^2 - R^2}{2dr}\right) \\ &\quad + R^2 \cos^{-1}\left(\frac{d^2 + R^2 - r^2}{2dR}\right) \\ &\quad - \frac{1}{2}\sqrt{\begin{array}{c}(-d+r+R) \times (d+r-R) \times (d-r+R) \\ \times (d+r+R)\end{array}}. \end{aligned}$$

where,

- R is the radius of circle number one, which is the area of Bombardier Beetle;
- r is the radius of circle number two, which is the area of predator;

- d is the distance between the centers of the circles.

As we aforementioned, mathematically speaking, we have a set of possible solutions, which are as follows:

1. If d is greater than the radius of circle number one and two, this means that The circles do not overlap, which the intersection area is equal to zero.
2. If d less than or equal to absolute value of subtraction between the radius of circle number one and two, this means that one circle is fully inside the other.

The Bombardier Beetle select the position of predator to activate the defense mechanism. Depends on the modeling of Bombardier Beetle movement and intersection area between Bombardier Beetle and the predator, a new position for each individual of the population should be calculated using Eq. (31). Hence, the procedure will replace the prior location if the new objective function value is better than the old one using Eq. (32).

$$x_{i,d}^{PBB} = \frac{\left(x_{i,d} + \left(SelectePredator_{i,d} \times CIA \times CR \cdot x_{i,d}\right)\right)}{Spray} \quad (31)$$

, $i = 1,2,\dots,N$, and $d = 1,2,\dots,m$

$$X_i = \begin{cases} X_i^{P1}, & F_i^{Obj} < F_i \\ X_i, & \text{else} \end{cases} \quad (32)$$

$$Spray = \left(chaos \times 2.7^{\left(100^{\left(\frac{iteration}{Max\_iterations}\right)}\right)}\right) \quad (33)$$

where,

- $x_{i,d}^{PBB}$ is the new suggested position of i[th] Bombardier Beetle in the d[th] dimension;
- $SelectePredator_{i,d}$ denotes the position while the d[th] is idimension of the predator;
- CIA is the circle intersection area between Bombardier Beetle position and predator position. This value can be calculated using Eq. (30). There are three possible solutions, first no intersection between two areas, which means the area is zero. Second, there is an intersection point between two areas. Third, the area between Bombardier Beetle position and predator position are overlapped;
- CR is the chemical reaction of toxic spray, which contains the following components:
  o hot water vapor, which is a value of 100. This value is boiling value of water;
  o oxygen value, which is random value between [0, 1];
  o irritating substances like p-benzoquinones, which is random value between [0, 1].
- Spray is the exponential equation of spreading the chemical reaction of toxic spray, which can be calculated using Eq. (33).
  o Where, chaos is used to simulate the real chaos of spraying the toxic spray over the problem-solving space domain, which includes the sinusoidal map, Chebyshev map, circle map, singer map, Gauss/mouse map, tent map, and iterative map [5];
  o Iteration is the iteration value;
  o Max iteration is the maximum iteration of procedure.
- N is the number of Bombardier Beetle;
- m is the number of decision variables.
- $F_i^{Obj}$ is the objective function value.

### c. Newton's Explanation of Insect Lift

Newton's second and third laws of motion, which serve as the foundation for his explanation of insect lift [22]:

- Second Law of Newton is:
  o Force is equal to mass multiplied by acceleration.
- Third Law of Newton is:
  o Action is equal to reaction.

An upward reactive force, or lift, is produced when an insect flaps its wings, accelerating air downward as depicted in Fig.9. The empirical lift equation for flapping flight can be formulated in the equation (34) [22]. This is similar to airplane aerodynamics but modified for unsteady flapping motion. The full procedure of Bombardier Beetle Optimizer (BBO) is listed in Algorithm 1.

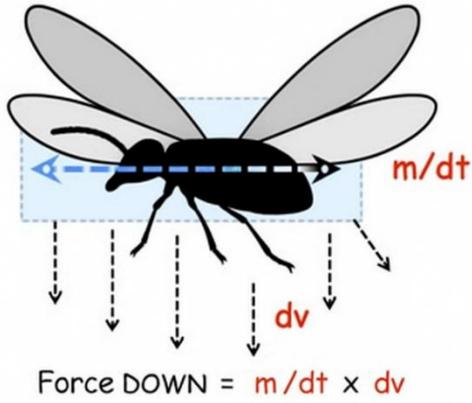

**Fig 9.** Flapping flight, an upward reactive force

$$L = LC * 0.5 * \rho * V^2 * A \quad (34)$$

where,
- LC is lift coefficient (experimentally determined for insect wing motion);
- $\rho$ is air density;
- V is wing's relative velocity;
- A is the effective wing area.

---

Algorithm 1. Pseudocode of BBO.

Start BBO.
1. Enter the variables, restrictions, and objective function that make up the issue.
2. Set the BBO maximum number of iterations (Max_Iter), and population size (N).
3. Using Equation (29), generate the initial population matrix
4. Evaluate the objective function.
5. For $t = 1$ to Max_Iter:
6. For $i = 1$ to $N$:
 7. Section # 1: defense mechanism (exploration)
 - Calculated the intersection area between Bombardier Beetle and predator locations using Equation (30).
 - Apply spray defense mechanism using Equation (33).
 - Calculate the new position of population and update the solution using Equation (31).
 8. Section #2: escape (insect left and fly away) (exploitation)
 - Calculate the left equation of Bombardier Beetle using Equation (34).
 - Calculate the new position of Bombardier Beetle using the following Equation:
$$\frac{L \times (ub_d - lb_d)}{t}$$
 - Calculate the new position of population and update the solution using Equation (31). Update fitness value using Equation (32).
9. end
10. Record the best candidate solution so far.
11. end
12. Output the optimal solution of the objective function.

End BBO.

---

## 4 Experimental and Result

In this section, CDO, SSO, GSA, PSO, BTO, GWO along with the proposed BBO are coded in ''MATLAB R2023a'' and run on Intel core i5 CPU with 6 GB RAM operating Windows 11. For every test of every algorithm, a total evaluation is repeated ten time runs to ensure the correctness and conversions of every algorithm. The performance of the proposed ''Bombardier Beetle Optimizer (BBO)'' is estimated using the well-known twenty-three benchmark functions of ''Congress on Evolutionary Computation (CEC) 2017" [4] along with all other algorithms. These test bed suites are minimization problems, which the best fitness (optimal value) of functions from 1 to 12 expect function 8 are zeros. The best fitness of function 8 is - 12,569.5. On the other hand, the best fitness of functions from 13 to 23 are ''– 1.15044, 0.998, 0.0003075, - 1.0316, 0.398, 3, - 3.86, - 3.32, - 10.2, - 10.4, - 10.5, and - 12,569.5''. These bed test problems are formulated in Appendix A.

The quality of results takes in mind three main things for qualitative test, which are mean ($\mu$), standard deviation ($\sigma$), and best fitness (optimal value), which the best fitness values are recorded at the end of each iteration for every algorithm. The proposed algorithm BBO and the other algorithms have various parameters in which are configured at the beginning of the procedures. These parameters are listed in the Table 2. These algorithms are compared using the following statistics:

- Mean ($\mu$): The following formula can be used to determine the average fitness values after a certain number of generations T:

$$Mean(\mu) = \frac{\sum_{t=1}^{n}(f_t)}{T}, \quad (35)$$

- Standard deviation ($\sigma$): Eq. (36) can be used to determine the standard deviation between populations after a certain number of generations T. This component represents the variation in the objective model's values.

$$Std(\sigma) = \sqrt{\frac{t}{T-1}\sum_{t=1}^{n}(f_t - mean)^2}, \quad (36)$$

- Optimal value (best fitness): The following (37)

formula can be used to get the minimum fitness value after a certain number of generations T:

$$Best = \min_{1 \leq i \leq T} f_i$$

**Table 2** Parameters of the algorithms

| Parameters | Value |
|---|---|
| **SSO** | |
| Damping factor of velocity (D) | Rand (0, 1) |
| pH | Rand (7, 14) |
| Temperature | Rand (35.5, 38.5) |
| Size of population (swarm size) | 30 |
| Numbers of iterations/generations | 1000 |
| **GSA** | |
| α | 20 |
| $G_0$ | 1 |
| Size of population | 30 |
| Numbers of iterations/generations | 1000 |
| **CDO** | |
| $S_\gamma$ is the speed of gamma | Rand (1, 300,000) **km/s** |
| $S_\beta$ is the speed of beta | Rand (1, 270,000) **km/s** |
| $S_\alpha$ is the speed of alpha | Rand (1, 16,000) **km/s** |
| r is the radius of radiations propagation | Rand (0, 1) |
| Size of population | 30 |
| Numbers of iterations/generations | 1000 |
| **BTO** | |
| CUG | $6.67 \times 10^{-11} Nm^2 Kg^{-2}$ |
| $area_{min}$ | log(500000) |
| $area_{max}$ | log(1,510,000) |
| Size of population | 30 |
| Numbers of generations | 1000 |
| **GWO** | |
| Size of population | 30 |
| Numbers of iterations/generations | 1000 |
| **BBO** | |
| Distance between Bombardier Beetle and predator (d) | Rand (0, 1) |
| r1 and r2 radius of circle 1 and circle 2 | Rand (0, 1) |
| Typical insect wing lift coefficient (CL) | Rand (0, 1) |
| Air density (ρ) | Rand (0, 1) |
| Wing area (A) | Rand (0, 1) |
| Wing velocity (V) | Rand (0, 1) |
| hot water vapor of toxic spray | 100 |
| oxygen value | Rand (0, 1) |
| p-benzoquinones | Rand (0, 1) |

**Table 3** numerical and statistical results of test suites of BBO, GWO, BTO, PSO, GSA, SSO, and CDO

| | | CDO | SSO | GSA | PSO | BTO | GWO | BBO |
|---|---|---|---|---|---|---|---|---|
| F1 | Best | 2.29E-262 | 7.58E-228 | 1.01E-16 | 2.60E-10 | 0.00E+00 | 6.73E-60 | 0.00E+00 |
| | Mean | 2806.973 | 146.1002 | 211.3284 | 546.5455 | 6.237578 | 385.47 | 75.05441 |
| | Worst | 7.12E+04 | 64754.96 | 6.06E+04 | 7.86E+04 | 673.4686 | 7.89E+04 | 5.63E+04 |
| | Std | 9218.187 | 2452.828 | 2725.425 | 4805.066 | 43.35072 | 3768.091615 | 1838.36768 |
| | Rank | 2 | 3 | 5 | 6 | 1 | 4 | 1 |
| F2 | Best | 2.79E-135 | 1.24E-129 | 7.53E-08 | 6.52E-06 | 3.1371e-310 | 5.70E-35 | 0.00E+00 |
| | Mean | 2.14E+10 | 3.78E+08 | 11333683 | 6.9E+09 | 0.402896 | 8.82E+09 | 1.88E+09 |
| | Worst | 1.96E+13 | 3.78E+11 | 1.13E+10 | 6.90E+12 | 21.98549 | 8.82E+12 | 1.87E+12 |

|  |  | CDO | SSO | GSA | PSO | BTO | GWO | BBO |
|---|---|---|---|---|---|---|---|---|
|  | Std | 6.18E+11 | 1.2E+10 | 3.58E+08 | 2.18E+11 | 2.084922 | 2.78957E+11 | 59331968562 |
|  | Rank | 3 | 4 | 6 | 7 | 2 | 5 | 1 |
|  | Best | 1.83E-226 | 5.68E-111 | 5.79E+02 | 13.89951 | 0.00E+00 | 7.54E-15 | 0.00E+00 |
|  | Mean | 85807.05 | 486.0306 | 1025.112 | 1474.157 | 307.9632 | 1706.965 | 205.8413 |
| F3 | Worst | 2.05E+05 | 76138.58 | 1.03E+05 | 8.60E+04 | 14244.81 | 1.76E+05 | 1.55E+05 |
|  | Std | 100057.6 | 4991.478 | 4559.733 | 6412.374 | 1536.95 | 9973.189196 | 5093.214413 |
|  | Rank | 2 | 3 | 6 | 5 | 1 | 4 | 1 |
|  | Best | 1.52E-126 | 1.14E-90 | 2.4705831 | 0.6105174 | 2.39E-303 | 1.58E-15 | 0.00E+00 |
|  | Mean | 11.00552 | 0.862714 | 3.718571 | 3.157701 | 7.365007 | 1.450779 | 0.167414 |
| F4 | Worst | 9.18E+01 | 87.73271 | 8.54E+01 | 8.92E+01 | 70.23451 | 8.68E+01 | 9.00E+01 |
|  | Std | 23.72986 | 6.270643 | 5.20553 | 7.254041 | 18.39352 | 7.920057621 | 3.286272631 |
|  | Rank | 3 | 4 | 7 | 6 | 2 | 5 | 1 |
|  | Best | 27.239300 | 28.084450 | 26.834915 | 22.63784 | 8.91678572 | 28.7275415 | 0.00E+00 |
|  | Mean | 7459448 | 679417.8 | 284805.2 | 482349.6 | 38370.17 | 666412.5 | 348228.7 |
| F5 | Worst | 1.72E+08 | 2.46E+08 | 2.40E+08 | 2.41E+08 | 15543596 | 2.61E+08 | 3.22E+08 |
|  | Std | 26860037 | 10241293 | 7657691 | 8810990 | 658704.6 | 10597955.6 | 10224107.88 |
|  | Rank | 5 | 6 | 4 | 3 | 2 | 7 | 1 |
|  | Best | 7.5 | 4.6204259 | 297.666 | 2.72E-09 | 0.46577979 | 0.75328863 | 0.00E+00 |
|  | Mean | 3585.788 | 111.1483 | 495.7094 | 567.6743 | 408.026 | 320.0206 | 97.522 |
| F6 | Worst | 6.34E+04 | 57081.7 | 7.55E+04 | 7.45E+04 | 9572.447 | 6.43E+04 | 7.07E+04 |
|  | Std | 11301.18 | 1959.556 | 3336.817 | 5015.746 | 1583.102 | 3318.004931 | 2335.554615 |
|  | Rank | 6 | 5 | 7 | 2 | 3 | 4 | 1 |
|  | Best | 3.19E-05 | 9.24E-06 | 0.0725943 | 0.043291 | 6.06E-05 | 0.00149316 | 9.15E-06 |
|  | Mean | 4.432104 | 0.267079 | 6.820644 | 34.42917 | 0.463912 | 0.392273 | 0.111119 |
| F7 | Worst | 1.21E+02 | 156.8493 | 1.33E+02 | 1.26E+02 | 6.40985 | 1.16E+02 | 110.5245 |
|  | Std | 16.00126 | 5.303608 | 22.82138 | 41.01692 | 1.488222 | 5.501127348 | 3.495081439 |
|  | Rank | 3 | 2 | 7 | 6 | 4 | 5 | 1 |
| Sum Rank |  | 24 | 27 | 42 | 35 | 15 | 34 | 7 |
| Mean Rank |  | 3.42 | 3.85 | 6 | 5 | 2.14 | 4.85 | 1 |

**Table 4** numerical and statistical results of test suites of BBO, GWO, BTO, PSO, GSA, SSO, and CDO

|  |  | CDO | SSO | GSA | PSO | BTO | GWO | BBO |
|---|---|---|---|---|---|---|---|---|
| F8 | Best | -3720.6698 | -5846.460 | -3415.703 | -5303.718 | -2056.1316 | -5315.5958 | -7650.9312 |
|  | Mean | -3655.94 | -5830.63 | -3412.61 | -4466.88 | -2029.17 | -3702.07 | -7152.78 |
|  | Worst | -2.57E+03 | -3592.65 | -1.84E+03 | -1.85E+03 | -1062.79 | -2.09E+03 | -1.66E+03 |
|  | Std | 109.3767 | 113.964 | 61.87957 | 1065.356 | 101.5579 | 896.9446982 | 1456.748813 |
|  | Rank | 5 | 2 | 6 | 4 | 7 | 3 | 1 |
| F9 | Best | 0.00E+00 | 0.00E+00 | 42.78318 | 55.87139 | 0.00E+00 | 0.00E+00 | 0.00E+00 |
|  | Mean | 181.6997 | 3.454446 | 68.88327 | 177.2212 | 0.161237 | 15.45727 | 1.039771 |
|  | Worst | 4.35E+02 | 456.0621 | 4.39E+02 | 4.75E+02 | 50.47899 | 4.57E+02 | 3.97E+02 |
|  | Std | 132.4407 | 32.17159 | 63.1026 | 116.5613 | 2.128815 | 57.47612808 | 18.31290715 |
|  | Rank | 1 | 1 | 2 | 3 | 1 | 1 | 1 |
| F10 | Best | 4.44E-15 | 8.88E-16 | 7.59E-09 | 3.04E-05 | 4.44E-16 | 1.87E-14 | 4.44E-16 |
|  | Mean | 2.785804 | 0.205432 | 0.630983 | 2.884515 | 3.454814 | 0.421708 | 0.064045 |
|  | Worst | 2.06E+01 | 20.50116 | 2.09E+01 | 2.07E+01 | 19.96677 | 2.04E+01 | 2.09E+01 |
|  | Std | 6.369129 | 1.46341 | 1.850071 | 3.373642 | 6.567251 | 2.306379999 | 1.032243665 |
|  | Rank | 3 | 2 | 5 | 6 | 1 | 4 | 1 |
| F11 | Best | 0.00E+00 | 0.00E+00 | 6.3328787 | 3.13E-10 | 0.00E+00 | 0.00815731 | 0.00E+00 |
|  | Mean | 36.014 | 1.827366 | 14.35131 | 24.33951 | 10.58792 | 3.450238 | 0.991813 |
|  | Worst | 6.52E+02 | 640.3442 | 6.04E+02 | 6.46E+02 | 120.942 | 6.76E+02 | 4.93E+02 |
|  | Std | 110.5404 | 25.06437 | 49.52972 | 94.12514 | 31.13853 | 37.48870613 | 17.57963583 |
|  | Rank | 1 | 1 | 4 | 2 | 1 | 3 | 1 |
| F12 | Best | 1.10610243 | 0.54128459 | 0.7488053 | 4.09E-10 | 1.42830782 | 0.0508685 | 0.72052502 |
|  | Mean | 13345615 | 1894428 | 787044.3 | 1282232 | 1.02E+08 | 1993671 | 636867.4 |
|  | Worst | 6.30E+08 | 5.81E+08 | 6.14E+08 | 6.51E+08 | 9.09E+08 | 4.94E+08 | 6.25E+08 |
|  | Std | 53182173 | 30251324 | 20086708 | 24188841 | 2.76E+08 | 27222262.55 | 19767131.11 |
|  | Rank | 6 | 3 | 5 | 1 | 7 | 2 | 4 |
| F13 | Best | 0.29475594 | 2.47406254 | 0.0109873 | 9.80E-13 | 0.37245674 | 0.60800782 | 1.35E-32 |
|  | Mean | 2.13E+07 | 3983608 | 1637630 | 1976444 | 1791279 | 4737926 | 1242062 |
|  | Worst | 1.17E+09 | 1.04E+09 | 1.22E+09 | 7.83E+08 | 35454329 | 1.46E+09 | 1.24E+09 |
|  | Std | 86557208.63 | 59904860 | 40544592 | 34762905 | 6579388 | 72326879.62 | 39277450.14 |
|  | Rank | 4 | 7 | 3 | 2 | 5 | 6 | 1 |
| Sum Rank | | 20 | 16 | 25 | 18 | 22 | 19 | 9 |
| Mean Rank | | 3.33 | 2.66 | 4.16 | 3 | 3.66 | 3.16 | 1.5 |

**Table 5** numerical and statistical results of test suites of BBO, GWO, BTO, PSO, GSA, SSO, and CDO

|     |       | CDO         | SSO        | GSA        | PSO        | BTO         | GWO         | BBO         |
|-----|-------|-------------|------------|------------|------------|-------------|-------------|-------------|
| F14 | Best  | 2.98210731  | 0.99800    | 1.9951324  | 1.992030   | 1.05809603  | 0.99800     | 0.99800     |
|     | Mean  | 5.422027    | 1.046626   | 2.430277   | 2.483369   | 1.277868    | 1.006784    | 1.089812    |
|     | Worst | 5.51E+00    | 9.96010    | 4.31E+02   | 2.33E+02   | 147.8734    | 1.27E+00    | 7.05E+01    |
|     | Std   | 0.234957768 | 0.451234   | 13.57987   | 7.67653    | 4.675419    | 0.046468991 | 2.223882028 |
|     | Rank  | 5           | 1          | 4          | 3          | 2           | 1           | 1           |
| F15 | Best  | 3.2888E-4   | 5.275E-4   | 2.3064E-3  | 9.823E-4   | 4.7342E-4   | 3.3766E-4   | 3.301E-4    |
|     | Mean  | 0.001275    | 0.001419   | 0.006211   | 0.001508   | 0.008319    | 0.000702    | 0.000597    |
|     | Worst | 7.93E-01    | 2.21E-01   | 1.41E-01   | 1.15E-01   | 0.065555    | 4.49E-02    | 1.71E-01    |
|     | Std   | 0.025089272 | 0.010285   | 0.011477   | 0.004655   | 0.019413    | 0.002455765 | 0.005886779 |
|     | Rank  | 2           | 5          | 7          | 6          | 4           | 3           | 1           |
| F16 | Best  | -1.03       | -1.03      | -1.03      | -1.03      | -1.03       | -1.03       | -1.03       |
|     | Mean  | -0.99329    | -1.02678   | -1.01131   | -1.03038   | -0.93465    | -1.03099    | -1.02763    |
|     | Worst | 2.13E+00    | 6.59E-01   | -1.92E-01  | -3.85E-01  | 1.274865    | -8.43E-01   | 6.81E-01    |
|     | Std   | 0.102033521 | 0.07725    | 0.088165   | 0.021135   | 0.336537    | 0.010347582 | 0.056393949 |
|     | Rank  | 1           | 1          | 1          | 1          | 1           | 1           | 1           |
| F17 | Best  | 0.39        | 0.39       | 0.39       | 0.39       | 0.39        | 0.39        | 0.39        |
|     | Mean  | 0.40501     | 0.410069   | 0.404563   | 0.400045   | 0.4222      | 0.398966    | 0.405971    |
|     | Worst | 1.93E+00    | 5.83E+00   | 7.06E-01   | 9.55E-01   | 1.08662     | 4.41E-01    | 1.35E+00    |
|     | Std   | 0.088137775 | 0.186291   | 0.043303   | 0.030564   | 0.097661    | 0.004661725 | 0.037003482 |
|     | Rank  | 1           | 1          | 1          | 1          | 1           | 1           | 1           |
| F18 | Best  | 3.0         | 3.0        | 3.0        | 3.0        | 3.0         | 3.0         | 3.0         |
|     | Mean  | 5.868765    | 3.297782   | 3.589883   | 3.831249   | 3.835106    | 3.067203    | 3.714141    |
|     | Worst | 8.09E+00    | 1.47E+02   | 3.62E+01   | 1.25E+02   | 34.90475    | 4.04E+01    | 9.53E+01    |
|     | Std   | 2.496927259 | 4.947175   | 2.782326   | 9.085635   | 4.694067    | 1.383289532 | 2.973006373 |
|     | Rank  | 1           | 1          | 1          | 1          | 1           | 1           | 1           |
| F19 | Best  | -3.86       | -3.81      | -3.86      | -3.86      | -3.80       | -3.86       | -3.83       |
|     | Mean  | -3.85422    | -3.81046   | -3.84975   | -3.86155   | -3.7249     | -3.85437    | -3.82974    |
|     | Worst | -3.70E+00   | -3.60E+00  | -2.01E+00  | -3.46E+00  | -2.05713    | -1.85E+00   | -3.03E+00   |
|     | Std   | 0.019301841 | 0.022962   | 0.075046   | 0.014402   | 0.315734    | 0.07746745  | 0.031540215 |
|     | Rank  | 1           | 3          | 1          | 1          | 4           | 1           | 2           |
| F20 | Best  | -3.28       | -2.99      | -3.32      | -3.32      | -2.981      | -3.201      | -3.137      |
|   | Mean  | -3.2623 | -2.8257 | -3.23636 | -3.18745 | -2.9269 | -3.12821 | -3.11651 |
|   | Worst | -1.90E+00 | -1.36E+00 | -7.18E-01 | -1.43E+00 | -1.34824 | -1.46E+00 | -1.85E+00 |
|   | Std   | 0.112827492 | 0.302575 | 0.207765 | 0.262526 | 0.266177 | 0.098696756 | 0.096196128 |
|   | Rank  | 2 | 6 | 1 | 1 | 5 | 3 | 4 |
| F21 | Best  | -9.3364808 | -4.1173383 | -10.153 | -10.153 | -3.784962 | -5.0551894 | -10.153 |
|   | Mean  | -8.43168 | -3.7516 | -9.13804 | -9.37367 | -3.19894 | -4.92772 | -10.074 |
|   | Worst | -4.94E-01 | -4.18E-01 | -3.05E-01 | -7.50E-01 | -0.88702 | -4.36E-01 | -3.68E-01 |
|   | Std   | 1.93810293 | 0.876071 | 2.654223 | 1.976018 | 1.133138 | 0.350718081 | 0.804617894 |
|   | Rank  | 2 | 4 | 1 | 1 | 5 | 3 | 1 |
| F22 | Best  | -7.94 | -3.21 | -10.40 | -5.122 | -4.52 | -10.40 | -10.40 |
|   | Mean  | -7.38868 | -2.94225 | -9.3969 | -4.85378 | -4.0088 | -7.8828 | -10.2733 |
|   | Worst | -3.77E-01 | -5.62E-01 | -3.26E-01 | -5.40E-01 | -0.7821 | -4.15E-01 | -5.42E-01 |
|   | Std   | 1.49386918 | 0.413207 | 2.65385 | 0.764644 | 1.242694 | 2.278754496 | 0.825878246 |
|   | Rank  | 2 | 5 | 1 | 3 | 4 | 1 | 1 |
| F23 | Best  | -8.657743 | -4.2703489 | -10.5 | -10.5 | -4.0144968 | -10.5 | -10.5 |
|   | Mean  | -7.59911 | -3.79037 | -9.37715 | -9.22537 | -3.46273 | -7.96162 | -10.3886 |
|   | Worst | -5.10E-01 | -1.33E+00 | -8.18E-01 | -4.97E-01 | -0.81296 | -6.90E-01 | -5.47E-01 |
|   | Std   | 1.408646007 | 0.937543 | 2.793826 | 2.848171 | 1.146176 | 2.379579853 | 0.907720507 |
|   | Rank  | 2 | 3 | 1 | 1 | 4 | 1 | 1 |
| Sum Rank | | 19 | 30 | 19 | 19 | 31 | 16 | 14 |
| Mean Rank | | 1.9 | 3 | 1.9 | 1.9 | 3.1 | 1.6 | 1.4 |

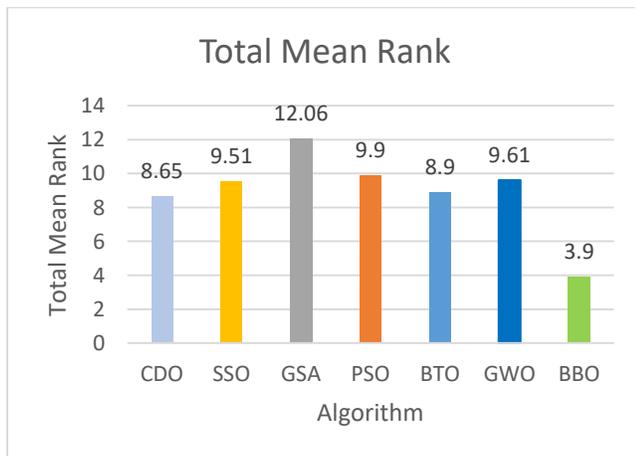

**Fig 10.** Mean rank of algorithms

In prior tables from 3 to 5 of evaluating CEC2017 test suites, we can notice that the best fitness has a highlighted background. Ten-times run are conducted to evaluate the optimization algorithms. Statistically speaking, BBO is the best on twenty benchmark function, which are f1, f2, f3, f4, f5, f6, f7, f8, f9, f10, f11, f13, f14, f15, f16, f17, f18, f21, f22, and f23. BBO gets the exact optimal value of these benchmark functions, which is superior and faster than the other algorithms in reaching the optimal value especially in solving the rugged and noisy problems, such as problems from one to eleven. This is clear in the figures of convergence of these benchmark functions. On the other hand, CDO is the best on seven benchmark functions, which are f9, f11, f16, f17, f18, f19, and f20. BTO is the best on eight benchmark functions, which are f1, f3, f9, f10, f11, f16, f17, and f18. SSO is the best on six benchmark functions, which are f9, f11, f14, f16, f17, and f18. The BTO can obtain the global minima as CDO, which the all have narrow domains. GWO is the best on eight benchmark functions, which are f9, f14, f16, f17, f18, f19, f22, and f23. SSO is the best on six benchmark functions, which are f9, f11, f14, f16, f17, and f18. PSO is the best on six functions, which are f12, f16, f17, f18, f21, and f23. GSA is the best on seven benchmark functions, which are f16, f17, f18, f19, f21, f22, and f23. However, for functions 9, 11, and 16, SSO and CDO perform in the same rank. BBO and BTO beat in the same rank in solving f1, f3, f9, f11, f14, and f22. On the other hand, all the algorithms beat in the same rank in solving f16, f17, and f18.

Fig. 10 shows the comparisons between BBO algorithm against the other algorithms in term of total mean rank in optimizing the twenty-three benchmark functions of CEC2017, which the algorithm that has the lowest mean rank is the best one. The BBO comes first with lowest mean rank of 3.9. This is clear in the convergence rates and experimental results of the BBO method as depicted in the convergence figures of test suites of f1, f2, f3, f4, f5, f6, f7, f8, f9, f10, f11, f13, f14, f15, f16, f17, f18, f21, f22, and f23 in which BBO is faster and superior of reaching and obtaining the global minima of these test beds than the other approaches. On the other hand, CDO comes in second rank, and BTO comes in the third rank. In addition, SSO comes in the fourth rank with total mean rank equal to 9.51. GWO comes in the fifth rank followed by PSO and GSA.

## 5 Discussion

BTO, CDO and GSA are very powerful physical based algorithms. However, they facing many problems, some of which BTO is mainly has good results in solving unimodal, but has slow convergence in solving multimodal problems. CDO tendency to fall into local minima. In addition, GSA is very slow, which its ability to reach the global minima is not good. This means that exploitation mechanism of GSA is not huge. In different view, PSO, SSO and GWO are swarm based methods, which they can be slightly fallen in the local minima, but they have precise and huge ability of exploitation. Depends on the prior mentioned weaknesses of these algorithms, we gain the chance to propose new optimization algorithm, called Bombardier Beetle Optimizer (BBO). Test bed problems of well-known CEC 2017 are utilized to compare the BBO against other well-known categories of optimization algorithms, such as PSO, SSO, GSA, GWO, CDO and BTO. Both numerical and statistical experimental are calculated for quantitative tests, which are best value (best fitness), mean rank, standard deviation ($\sigma$), and mean ($\mu$). Furthermore, the rate of algorithm convergence has been drawn for each algorithm of each benchmark function for qualitative test. The iteration number for each algorithm is set to one thousand. These tests are repeated ten-time runs to ensure the convergence of results for each benchmark function.

Overall, the proposed BBO algorithm outperforms all the other algorithms on the majority of the test bed suites of CEC2017, which has merit in exploration mechanisms in solving the narrow and wide domain of these test bed problems. This is very clear in the previous tables from 3 to 5 of results and the rate of algorithm convergence in figures of test bed problems of f5, f6, f7, f8, f9, f10, f11, f13, f14, f15, f16, f17, f18, f21, f22, and f23, and mainly for rugged and noisy problems, such as test beds from one to four. Those evinced the superiority of the proposed BBO in term of convergence in comparison with the other algorithms to export the problem domain.

The test suites that are used in this work might have limitations, and different tuning parameters could be accessible in actual studies to alter the results. To guarantee its efficacy, the suggested BBO should be put to the test in solving of actual real life problem in the future. Examples of these problems are maximizing signal propagation in "Radio Networks (RNs), Under Water Sensors" and minimizing attenuation and interference factors in "Wireless Sensor Networks (WSN)" [23–27].

## 6 Conclusion

In this article, a new optimization algorithm, called "Bombardier Beetle Optimizer (BBO)" has been proposed. The main principles of this approach is inspired by the defense mechanism of Bombardier Beetle against the predators, which the Bombardier Beetle triggers a toxic chemical spray when it feels threatened. Twenty-three well-known test problems of CEC2017 have been chosen to compare the BBO against other well-known algorithms, such as PSO, SSO, GSA, GWO, BTO, and CDO. These approaches are compared using two methodologies, known as qualitative and quantitative tests. The mean ($\mu$), standard deviation ($\sigma$), and optimal value (best fitness) are computed for the quantitative technique. Conversely, for the qualitative technique's the convergence rate is determined by the results of each approach and is drowned for every test bed problem. From the results, we can notice that the proposed BBO has merit in exploration mechanisms in solving the narrow and wide domain of these test bed problems. This is very clear in the previous tables of results and the rate of algorithm convergence in figures of test bed problems of f5, f6, f7, f8, f9, f10, f11, f13, f14, f15, f16, f17, f18, f21, f22, and f23, and mainly for rugged and noisy problems, such as test beds from one to four. We can conclude that the BBO can be considered as a more efficient feasible alternative algorithm since it outperformed PSO, SSO, GSA, GWO, BTO, and CDO in most test bed problems of rugged and noisy domain of search space.

**Conflicts of Interest**

The authors declare that there are no conflicts of interest regarding the publication of this paper.

**Ethical Approval & Consent for publication:**

We give our consent for the publication of identifiable details, which can include photograph(s) and/or videos and/or case history and/or details within the text ("Material") to be published in the above Journal and Article. We confirm that we have seen and been given the opportunity to read both the Material and the Article (as attached) to be published by your journal. In Addition, a sample of data of this paper will be available upon request. The open source code of our algorithms, namely, CDO, SSO and BTO are available via the following links:

**Chernobyl Disaster Optimizer (CDO):**

https://www.mathworks.com/matlabcentral/fileexchange/124351-chernobyl-disaster-optimizer-cdo

**Bermuda Triangle Optimizer (BTO):**
https://github.com/sh7adeh1990/BTO

**Sperm Swarm Optimization (SSO):**

https://www.mathworks.com/matlabcentral/fileexchange/92150-sperm-swarm-optimization-sso

**Appendix A**

"Congress on Evolutionary Computation (CEC 2017)" test bed suites are mathematically formulated in Table 6 [4].

Table 6 CEC 2017 test bed problems

| Function | Dim | Range |
|---|---|---|
| $F_1(z) = \sum_{i=1}^{d} z_i^2$ | 30 | [-100, 100] |
| $F_2(z) = \sum_{i=1}^{z} \|z_i\| + \prod_{i=1}^{d} \|z_i\|$ | 30 | [-10, 10] |
| $F_3(z) = \sum_{i=1}^{d} \left( \sum_{j-1}^{i} z_j \right)^2$ | 30 | [-100, 100] |
| $F_4(z) = \max_i \{\|z_i\|, 1 \leq i \leq d\}$ | 30 | [-100, 100] |
| $F_5(z) = \sum_{i=1}^{d-1} \left[ 100(z_{i+1} - z_i^2)^2 + (z_i - 1)^2 \right]$ | 30 | [-30, 30] |
| $F_6(z) = \sum_{i=1}^{d} ([z_i + 0.5])^2$ | 30 | [-100, 100] |
| $F_7(z) = \sum_{i=1}^{d} i z_i^4 + random[0,1)$ | 30 | [-1.28, 1.28] |
| $F_8(z) = \sum_{i=1}^{d} -z \sin(\sqrt{\|z_i\|})$ | 30 | [-500, 500] |
| $F_9(z) = \sum_{i=1}^{d} [z_i^2 - 10\cos(2\pi z_i) + 10]$ | 30 | [-5.12, 5.12] |
| $F_{10}(z) = -20\exp\left(-0.2\sqrt{\frac{i}{z}\sum_{i=1}^{d} z_i^2}\right) - \exp\left(\frac{i}{z}\sum_{i=1}^{d}\cos(2\pi z_i)\right) + 20 + e$ | 30 | [-32, 32] |
| $F_{11}(z) = \frac{i}{4000}\sum_{i=1}^{d} z_i^2 - \prod_{i=1}^{d}\cos\left(\frac{z_i}{\sqrt{i}}\right) + 1$ | 30 | [-600, 600] |
| $F_{12}(z) = \frac{\pi}{d}\left\{10\sin(\pi z_1) + \sum_{i=1}^{d-1}(v_i - 1)^2[1 + 10\sin^2(\pi v_{i+1})] + (v_z - 1)^2\right\} + \sum_{i=1}^{z} u(z_i, 10, 100, 4)$ | 30 | [-50, 50] |
| $F_{13}(z) = 0.1\left\{ \sin^2(3\pi z_1) + \sum_{i=1}^{d}(z_i - 1)^2[1 + \sin^2(3\pi z_i + 1)] + (z_d - 1)^2[1 + \sin^2(2\pi z_d)] + \sum_{i=1}^{z} u(z_i, 5, 100, 4) \right\}$ | 30 | [-50, 50] |

| Function | Dim | Range |
|---|---|---|
| $F_{14}(z) = \left( \frac{1}{500} + \sum_{j=1}^{25} \frac{1}{j + \sum_{i=1}^{2} (z_i - a_{ij})^6} \right)^{-1}$ | 2 | [-62.536, 65.536] |
| $F_{15}(z) = \sum_{i=1}^{11} \left[ a_i - \frac{y_{1}(b_i^2 + b_i z_2)}{b_i^2 + b_i z_3 + z_4} \right]^2$ | 4 | [-5, 5] |
| $F_{16}(z) = 4z_1^2 - 2.1z_1^4 + \frac{1}{3}z_1^6 + z_1 z_2 - 4z_2^2 + 4z_2^4$ | 2 | [-5, 5] |
| $F_{17}(z) = \left( z_2 - \frac{5.1}{4\pi^2} z_1^2 + \frac{5}{\pi} z_1 - 6 \right)^2 + 10\left(1 - \frac{1}{8\pi}\right)\cos z_1 + 10$ | 2 | [-5, 5] |
| $F_{18}(z) = [1 + (z_1 + z_2 + 1)^2 (19 - 14z_1 + 3z_1^2 - 14z_2 + 6z_1 z_2 + 3z_2^2)] \times [30 + (2z_1 - 3z_2)^2 \times (18 - 32z_1 + 12z_1^2 + 48z_2 - 36z_1 z_2 + 27z_2^2)]$ | 2 | [-2,2] |
| $F_{19}(z) = -\sum_{i=1}^{4} c_i \exp\left( -\sum_{j=1}^{3} a_{ij}(z_j - p_{ij})^2 \right)$ | 3 | [1,3] |
| $F_{20}(z) = -\sum_{i=1}^{4} c_i \exp\left( -\sum_{j=1}^{6} a_{ij}(z_j - p_{ij})^2 \right)$ | 6 | [0,1] |
| $F_{21}(z) = -\sum_{i=1}^{5} \left[ (z - a_i)(z - a_i)^T + c_i \right]^{-1}$ | 4 | [0,10] |
| $F_{22}(z) = -\sum_{i=1}^{7} \left[ (z - a_i)(z - a_i)^T + c_i \right]^{-1}$ | 4 | [0,10] |
| $F_{23}(z) = -\sum_{i=1}^{10} \left[ (z - a_i)(z - a_i)^T + c_i \right]^{-1}$ | 4 | [0,10] |

## Authors' Information

**Dr. Hisham A. Shehadeh** received the B.S. degree in computer science from Al-Balqa` Applied University, Jordan in 2012, the M.S. degree in computer science from the Jordan University of Science and Technology, Irbid, Jordan, in 2014, and the Ph.D. degree from the Department of Computer System and Technology, University of Malaya (UM), Kuala Lumpur, Malaysia, in 2018. He was an Assistant professor of CS/AI in Amman Arab University from 2020 to 2024. He was a research assistant at UM from 2017 to2018. He was a Teaching Assistant and a lecturer with CS Department, College of Computer and Information Technology, Jordan University of Science and Technology from 2013 to 2014 and from 2014 to 2016 respectively. He was an Assistant Professor of Computer Science at the Department of Information Technology, Al-Huson University College, Al-Balqa Applied University, Al-Huson, Jordan from 9/2024 to 9/2025. Currently, he is an Assistant Professor of Computer Science at the Department of CS at YU, Irbid, Jordan.

**Prof. Mohd Yamani Idna Idris** has been a dedicated faculty member at Universiti Malaya since 2000, contributing significantly to education, research, and innovation. With interest and expertise in Optimization, Internet of Things (IoT), image processing, computer vision, and pattern recognition, his scholarly work has enriched both academic knowledge and practical applications. His extensive experience in teaching, learning, and research has influenced countless learners, and his research is widely published in reputed journals. As a mentor, he has successfully supervised 25 Ph.D. candidates, providing invaluable emotional and motivational support to ensure their success. His expertise is also sought after in academic circles, where he serves as a reviewer and editorial board member for journals like MJCS, Pattern Recognition, and IEEE Transactions on ITS. Additionally, he has been an external examiner for Ph.D. candidates and has contributed to international technical review panels. Beyond his research contributions, Dr. Idris has held key faculty leadership positions, including Head of Department and Deputy Dean (Undergraduate (acting) and Research & Development), playing a vital role in shaping academic direction and fostering innovation.

**Prof. Iqbal H. Jebril** is a Professor of Mathematics at Al-Zaytoonah University of Jordan, specializing in mathematical analysis, mathematical Optimization, fuzzy systems, and fractional calculus. He earned his Ph.D. in Mathematical Analysis from Universiti Kebangsaan Malaysia (2005). He has authored numerous Scopus-indexed publications and serves as Editor-in-Chief of IJASCA and other journals. His work bridges mathematics with engineering and computational sciences.

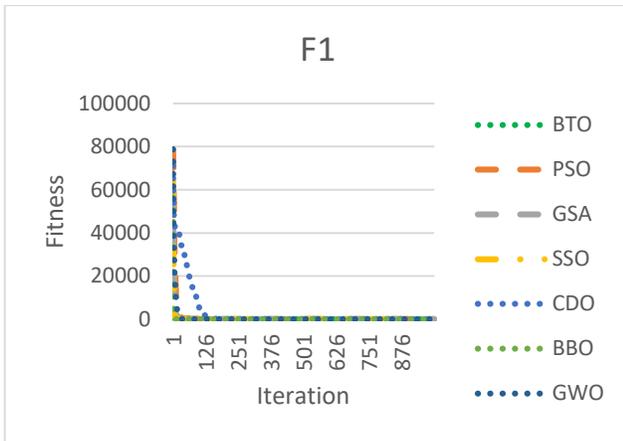

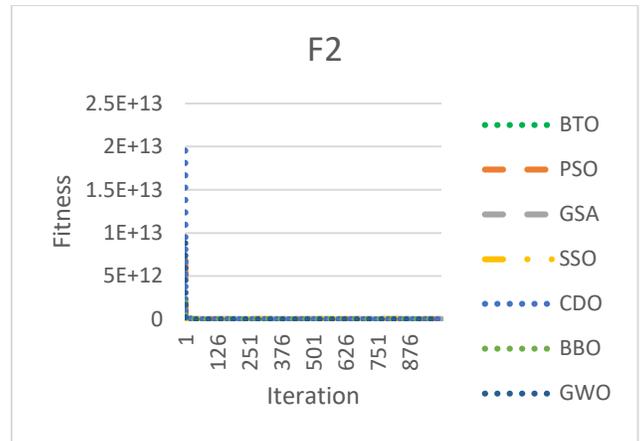

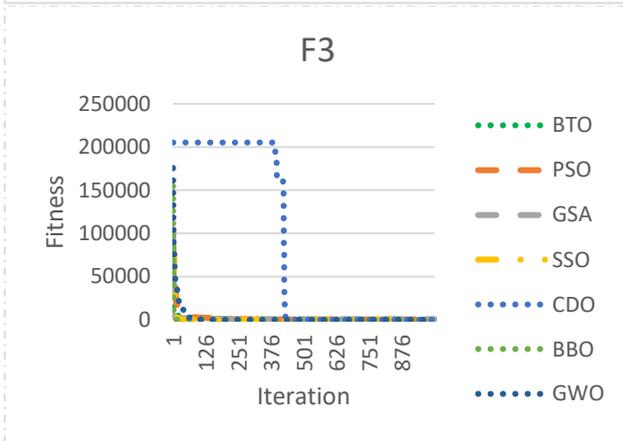

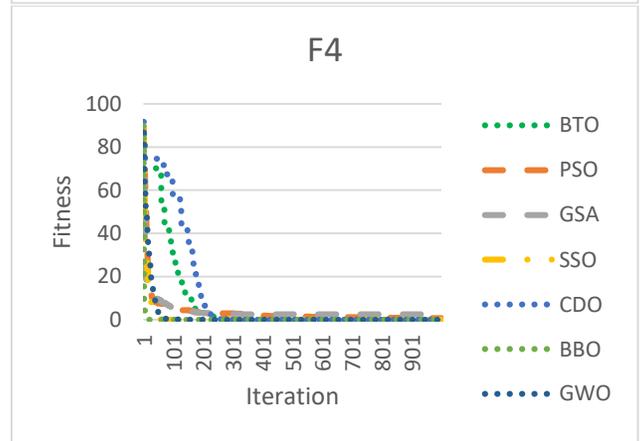

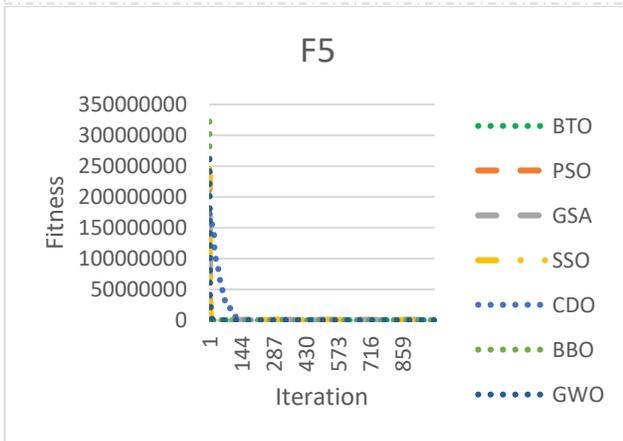

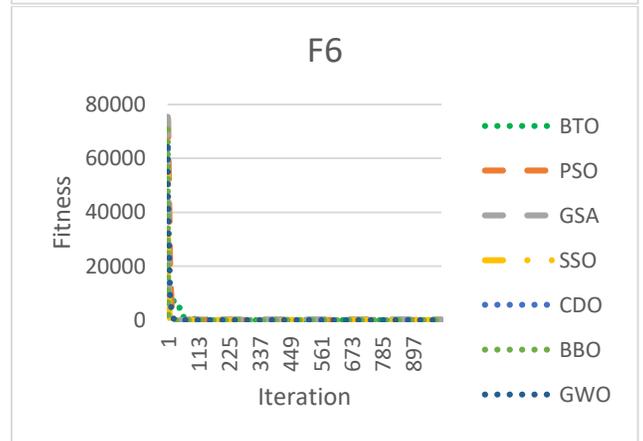

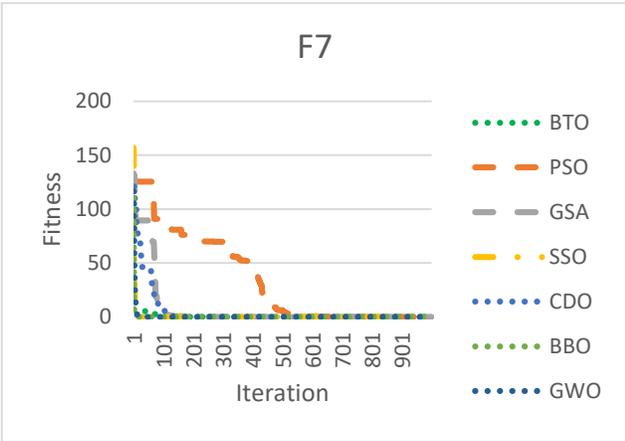
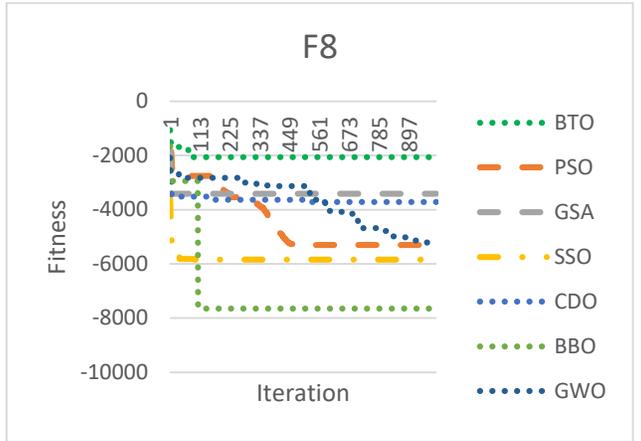
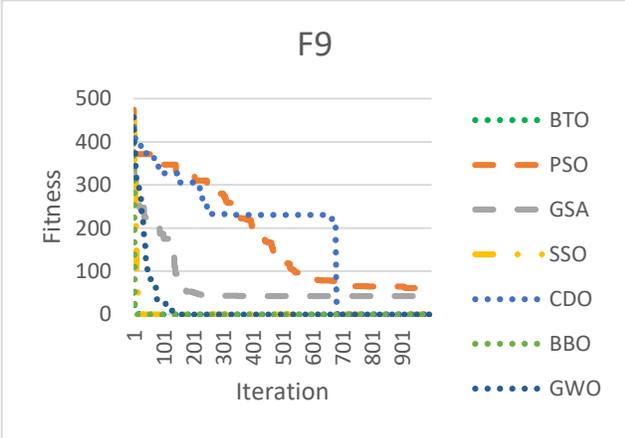
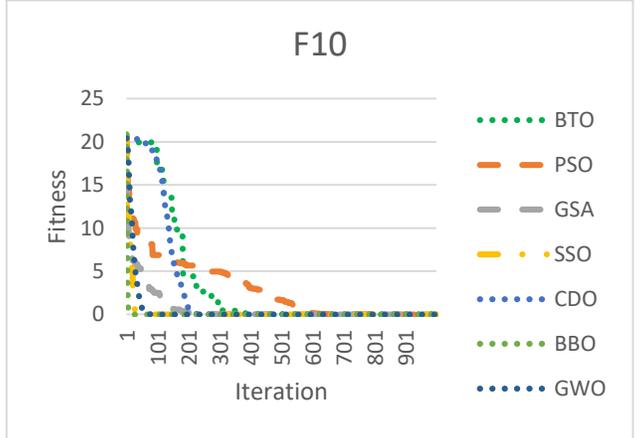
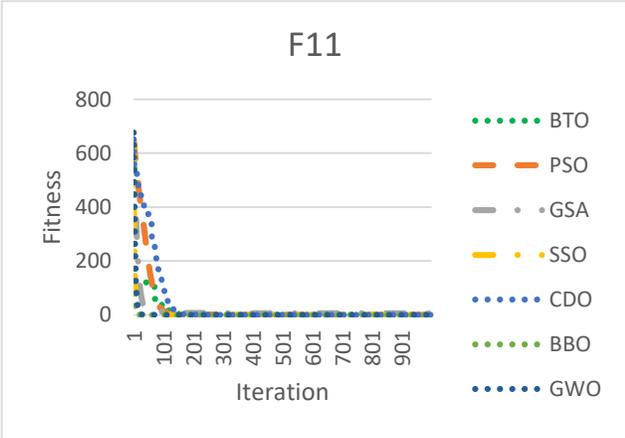
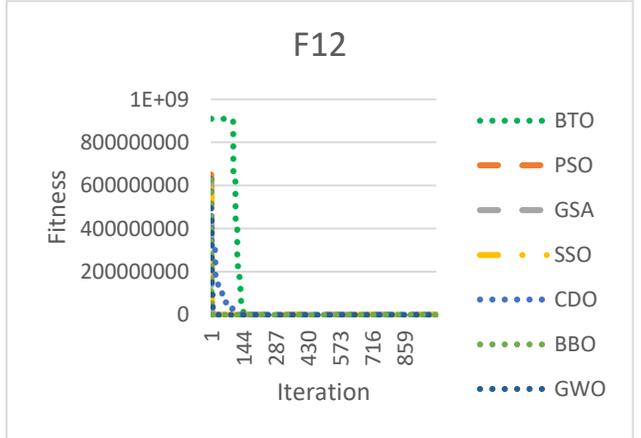
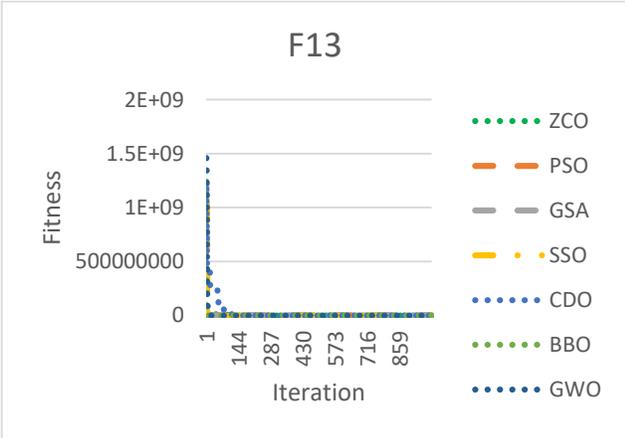
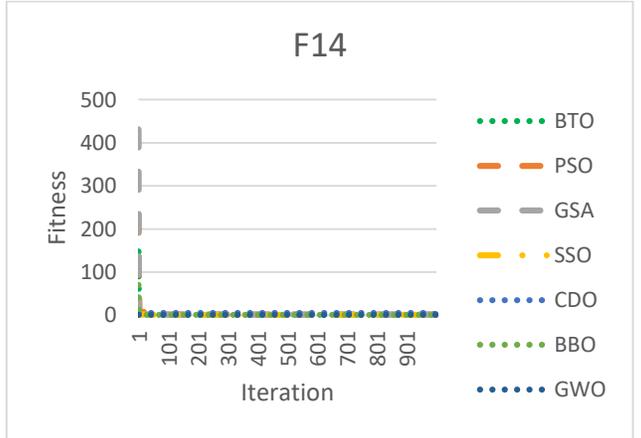

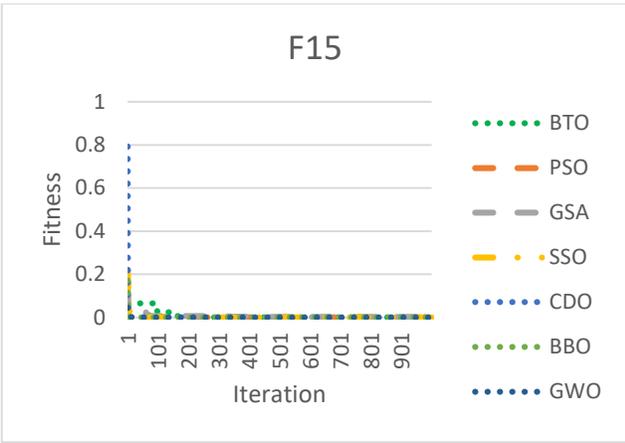
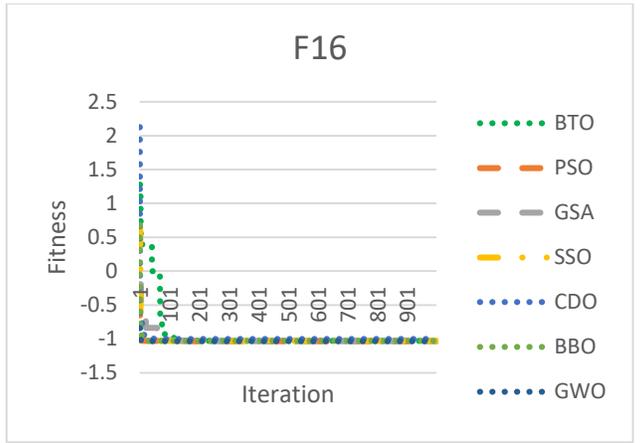
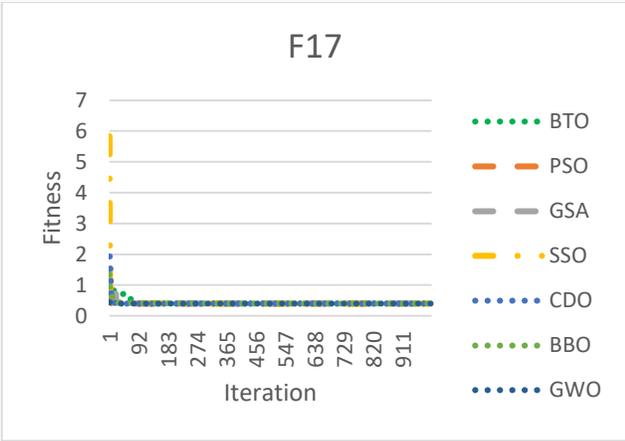
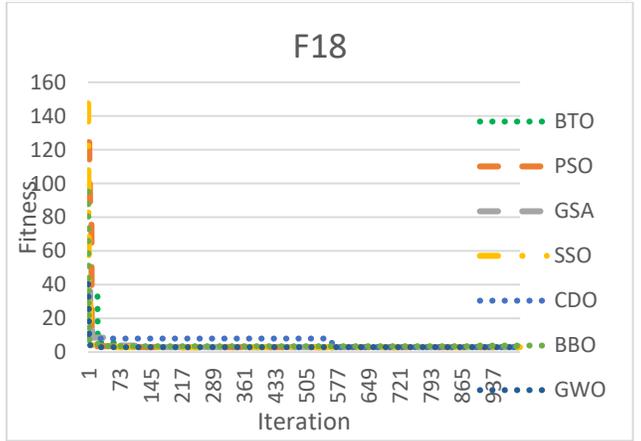
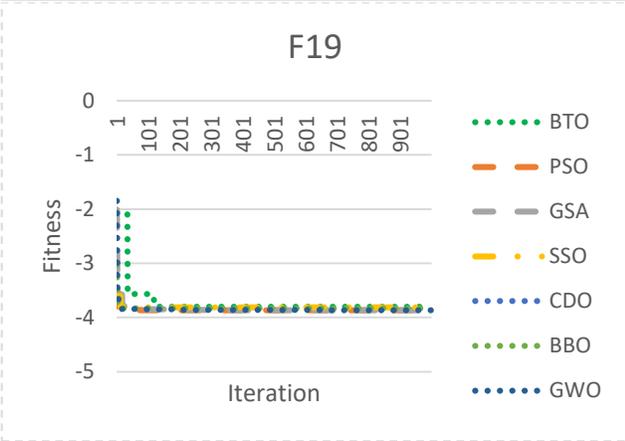
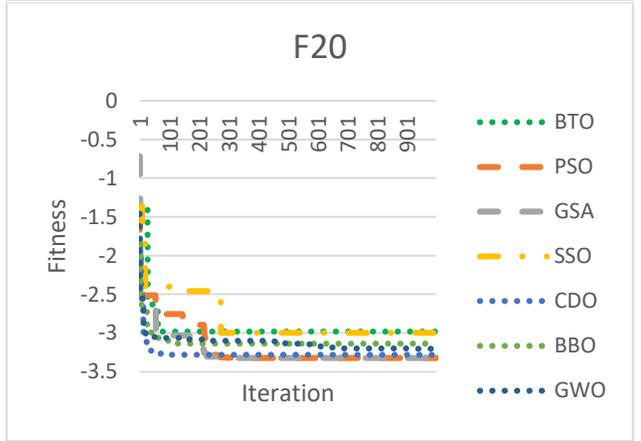
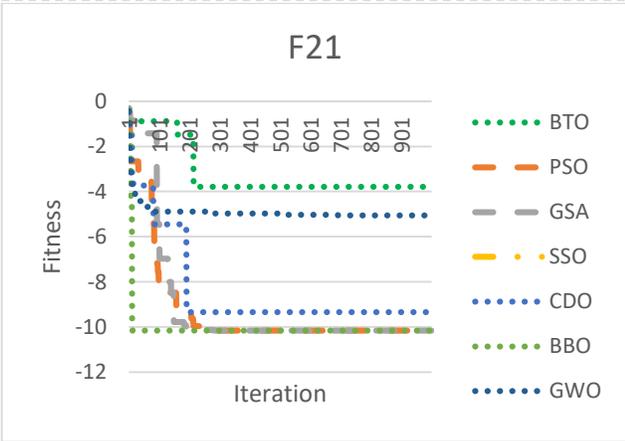
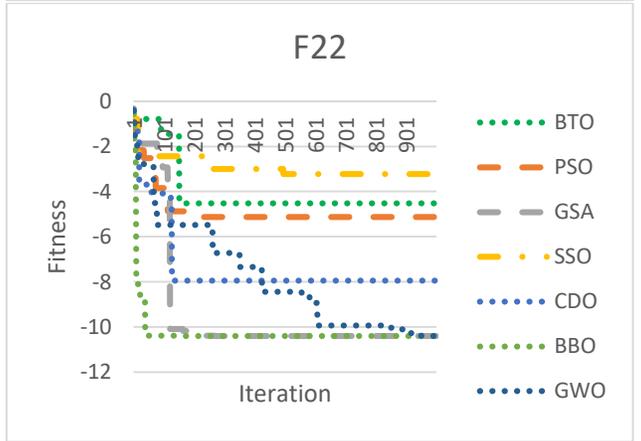

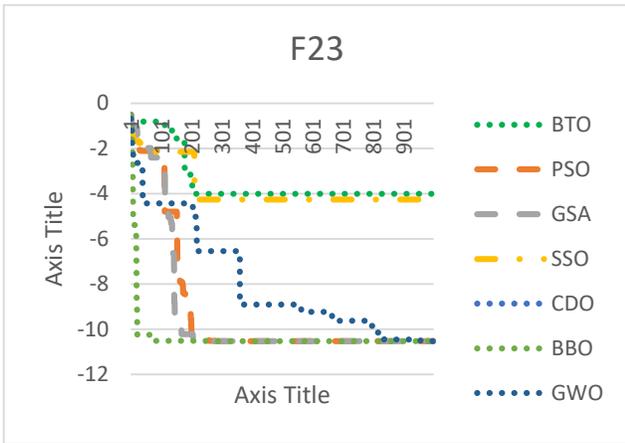

**Fig 11.** Convergence rate of the algorithms